\documentclass{article}
\PassOptionsToPackage{table}{xcolor}
\usepackage{iclr2027_conference,times}

\usepackage{amsmath,amsfonts,bm}

\def\eqref#1{equation~\ref{#1}}

\def\1{\bm{1}}

\DeclareMathAlphabet{\mathsfit}{\encodingdefault}{\sfdefault}{m}{sl}
\SetMathAlphabet{\mathsfit}{bold}{\encodingdefault}{\sfdefault}{bx}{n}

\usepackage{hyperref}
\usepackage{url}
\usepackage{amsmath,amssymb}
\usepackage{booktabs,adjustbox,graphicx,capt-of,xcolor,placeins}
\definecolor{oursblue}{RGB}{235,244,253}

\title{\raggedright
GeoState: Learning View-Decoupled \mbox{Geometry} States for 3D World Modeling
}

\author{\textbf{Hongchang Liu}$^{1}$, \textbf{Jian Liu}$^{1}$, \textbf{Yujuan Ding}$^{2}$, \textbf{Liang Peng}$^{3}$,\\
\textbf{Xun Jiang}$^{1}$, \textbf{Fei Ma}$^{4}$, \textbf{Xing Xu}$^{1}$, \textbf{Yi Bin}$^{1,\ast}$\\[0.5em]
$^{1}$Tongji University\\
$^{2}$Hong Kong Polytechnic University\\
$^{3}$Tencent\\
$^{4}$Guangdong Laboratory of Artificial Intelligence and Digital Economy (SZ)\\
{\fontencoding{T1}\selectfont\texttt{Hongchang\_Liu@outlook.com, yi.bin@hotmail.com}}
}

\iclrfinalcopy

\usepackage{placeins}

\begin{document}
\maketitle

\begin{abstract}
Recent 3D world models increasingly use features from geometry foundation models (GFMs) as predictive world states, benefiting from their rich geometric priors. However, the view dependence of these features entangles scene dynamics with camera-induced variation, making the resulting state evolution harder to predict. We propose GeoState, a view-decoupled geometry representation for predictive 3D world modeling that separates scene-state evolution from camera-dependent observation. Specifically, a scene tokenizer aggregates multi-view GFM features into a view-decoupled scene state, which a geometry decoder queries using camera rays. An autoregressive world model then predicts future scene states, while camera motion is modeled separately. The predicted states can therefore be decoded under either predicted or externally specified cameras, while retaining a single context-derived scale throughout the rollout without per-frame scale alignment. Experiments show that GeoState outperforms prior 3D world models on Waymo, supports geometrically consistent cross-view querying, and generalizes zero-shot to KITTI and nuScenes. \textit{Project
page:} \url{https://hongchang-liu.github.io/GeoState-project-page/}.
\end{abstract}

\section{Introduction}

Learning to predict how the physical world evolves is a central goal of world modeling~\citep{ha2018worldmodels}, with direct relevance to autonomous driving, robotics, and embodied intelligence. Visual world models predict the future in pixel space~\citep{hu2023gaia1,wang2023drivedreamer,gao2024vista,nvidia2025cosmos}, which can introduce substantial representational redundancy and make geometric consistency difficult to maintain.
This has motivated increasing interest in 3D world models, which represent and predict future environments in geometrically grounded state spaces. By explicitly encoding spatial structure and 3D relationships, these representations provide a structured and geometrically grounded state space that more directly reflects the real world.

Recent 3D world models have explored a range of structured geometric state spaces, including depth, occupancy, point-based representations, and 3D Gaussians~\citep{zheng2024occworld,zhang2024copilot4d,yang2024vidar,zhu2025aether,chen2025deepverse,zuo2025gaussianworld}. In parallel, geometry foundation models (GFMs) have substantially advanced feed-forward 3D perception by learning rich geometric representations directly from RGB observations~\citep{wang2024dust3r,yang2025fast3r,wang2025cut3r,wang2025vggt,wang2026vggtomega}. Benefiting from large-scale pretraining, these representations capture strong geometric priors that generalize across scenes and tasks, making them attractive state spaces for predictive world modeling. Building on this capability, recent approaches have begun to directly model temporal evolution in pretrained geometry feature spaces~\citep{sun2026vggtworld,lu2026geowam}. In particular, VGGT-World~\citep{sun2026vggtworld} treats frozen VGGT features as world states and predicts their future evolution, demonstrating the potential of GFM representations for 3D forecasting.

Despite the strong performance achieved by recent GFM-based 3D world models, a strong representation for geometric perception is not necessarily an appropriate state representation for world dynamics. GFM features are typically constructed from and organized around the observing views, so changes in camera viewpoint can induce substantial changes in the representation even when the underlying physical scene remains unchanged. When such features are directly used as predictive states, a temporal model would account simultaneously for genuine scene evolution and camera-induced variation, although these arise from fundamentally different sources. In particular, even a static scene can undergo substantial changes in feature space as the camera moves, forcing the temporal model to predict viewpoint-dependent transformations that do not correspond to physical scene evolution. This unnecessary variation makes the latent dynamics more complex and therefore harder to predict over long horizons. FR3D~\citep{morbitzer2026fr3d} also shows that separating ego-motion from scene forecasting can improve temporal modeling, but its predictive state still inherits the view-dependent structure of the underlying reconstruction representation. We argue that such separation should instead be built into the representation itself: \emph{the state should describe \textbf{what the world is}, while the camera should determine only \textbf{how that world is observed}}.

To this end, we introduce GeoState, a view-decoupled geometry state learned from pretrained GFM features. Given multi-view observations, a scene tokenizer compresses their geometry features into a compact scene state represented in a shared scene coordinate system, while the query viewpoint is introduced only during decoding. Specifically, a geometry decoder takes the scene state together with rays from a query camera and predicts the geometry observed from that viewpoint. Consequently, the same scene state can be queried repeatedly from different viewpoints without re-encoding the scene. This transforms view-conditioned GFM features into a representation in which scene geometry is explicitly separated from camera-dependent observation.

Building on this representation, we formulate 3D world modeling directly in the scene-state space. The temporal model can therefore focus solely on predicting scene dynamics, while camera evolution is modeled independently. At inference time, a predicted future state can be decoded using either a separately predicted camera trajectory or a user-controlled camera. Our formulation also maintains a consistent scene gauge throughout temporal prediction and geometry decoding, avoiding the need for per-frame scale alignment during rollout.

We evaluate our approach on large-scale dynamic driving scenes,
using Waymo~\citep{sun2020waymo} for in-domain evaluation and
KITTI~\citep{geiger2012kitti} and nuScenes~\citep{caesar2020nuscenes}
for zero-shot transfer. Our approach outperforms the compared world
models in depth forecasting across all reported horizons, while
maintaining a consistent scale throughout the forecasting rollout.
Experiments further demonstrate geometrically consistent querying
across viewpoints and generalization across driving datasets. Ablation
results show that view decoupling makes the scene representation more predictable. Together, these findings support view-decoupled geometry states derived from pretrained GFMs as an effective representation for predictive 3D world modeling.

Our contributions are threefold:
\begin{itemize}
    \item We identify viewpoint coupling in geometry foundation model
    features as a limitation when using them as predictive world
    states, and introduce \textbf{GeoState}, a view-decoupled geometry state
    derived from pretrained GFM representations that separates scene content
    from how the scene is observed.

\item We develop a scene-state learning framework that realizes GeoState through a Scene Tokenizer that aggregates multi-view GFM features into a shared scene state and a query-conditioned Geometry Decoder that introduces the query viewpoint only at decoding time.
\item We formulate a 3D world model that predicts scene evolution directly in GeoState space while modeling camera evolution separately, and demonstrate long-horizon geometry forecasting, geometrically consistent cross-view querying, and zero-shot generalization across driving datasets without per-frame scale alignment.
\end{itemize}

\section{Related Work}

\subsection{3D World Models}

Recent world models increasingly move beyond pixel-space prediction toward structured 3D scene modeling and forecasting. Copilot4D~\citep{zhang2024copilot4d} and ViDAR~\citep{yang2024vidar} forecast future point clouds, while OccWorld~\citep{zheng2024occworld} and DOME~\citep{gu2024dome} predict scene evolution in semantic occupancy space. DIO~\citep{diehl2025dio} represents future scene geometry and motion through decomposable implicit occupancy-flow fields, and GaussianWorld~\citep{zuo2025gaussianworld} models scene evolution in 3D Gaussian space for streaming occupancy prediction. More recently, several approaches directly use learned geometry representations as predictive world states. VGGT-World~\citep{sun2026vggtworld} forecasts future VGGT features, while GeoWAM~\citep{lu2026geowam} learns geometry-aware predictive states for autonomous driving. FR3D~\citep{morbitzer2026fr3d} predicts the latent state of a pretrained 3D reconstruction model while modeling ego-motion separately. In contrast, we remove viewpoint dependence from the geometry state itself and introduce viewpoint only as a query condition during decoding.

\subsection{Geometry Foundation Models}

Geometry foundation models provide strong pretrained priors for multi-view 3D perception. DUSt3R~\citep{wang2024dust3r} formulates reconstruction through pointmap prediction, while Fast3R~\citep{yang2025fast3r} scales joint reconstruction to larger collections of views. CUT3R~\citep{wang2025cut3r} introduces a persistent latent state for continuously integrating observations, and VGGT~\citep{wang2025vggt} unifies camera estimation, depth, point maps, and tracking in a feed-forward architecture. VGGT-$\Omega$~\citep{wang2026vggtomega} further strengthens this paradigm for dynamic 3D scenes, making it a natural geometry backbone for our setting. We build on its pretrained geometric features and reformulate them into a compact, view-decoupled scene-level representation better suited for temporal world modeling.
\subsection{Compact 3D Representations}

A related line of work compresses multi-view observations into compact scene representations. LVSM~\citep{jin2025lvsm} encodes input views into a fixed-length set of latent tokens for novel-view synthesis, while Surflo~\citep{guedon2026surflo} aggregates dense VGGT features into a fixed-size global latent representation for efficient 3D surface reconstruction. 3DRAE~\citep{wei20263drae} further learns a fixed-length, view-decoupled 3D latent space that can be queried from arbitrary cameras. These works demonstrate that multi-view observations can be represented without retaining a separate dense representation for each view. While they focus primarily on static scene reconstruction or novel-view synthesis, our work investigates whether such scene-level representations can serve as predictive states for temporal world modeling.

\section{Method}
\label{sec:method}
\subsection{Method Overview}
\label{sec:problem}

\begin{figure}[t]
\centering
\includegraphics[width=\linewidth]{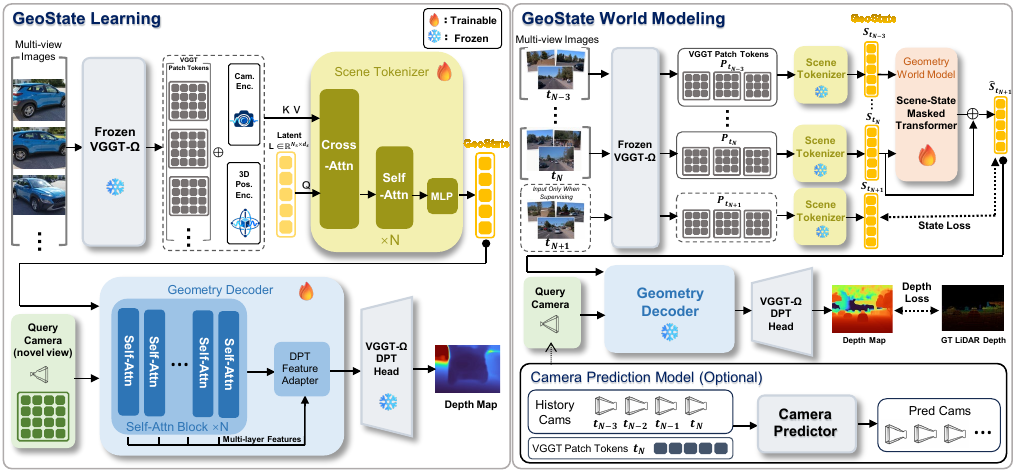}
\caption{\textbf{Overview of GeoState.}
GeoState Learning constructs view-decoupled scene states with a scene
tokenizer and query-conditioned geometry decoder. GeoState World Modeling
predicts future states directly in this representation space.
}
\label{fig:framework}
\end{figure}

Our goal is to predict future scene geometry without requiring the world
model to simultaneously account for variations induced by camera motion.
To this end, as illustrated in Figure~\ref{fig:framework}, we first construct a view-decoupled geometry state from
multi-view observations, and then model its temporal evolution while
handling camera evolution separately. 


\textbf{GeoState learning.}
Given multi-view RGB observations
$\mathcal{I}_t=\{I_t^v\}_{v\in\mathcal{V}_t}$, a frozen geometry
foundation model extracts geometry-aware features
$\mathbf{M}_t$, with the multi-view geometry expressed in a shared
scene coordinate system defined by a common anchor view. A learnable
Scene Tokenizer $E_{\phi}$ then aggregates $\mathbf{M}_t$ into a fixed-length
scene state:
\begin{equation}
    S_t = E_{\phi}(\mathbf{M}_t).
    \label{eq:scene_state}
\end{equation}
The state is not tied to any single observation viewpoint;
we refer to it as a \emph{view-decoupled geometry state}, or GeoState.
A faithful GeoState should capture the underlying 3D scene geometry,
which could be reflected in accurate and geometrically consistent
depth reconstruction across different viewpoints.
We therefore introduce a query-conditioned geometry decoder
that reconstructs a depth map from $S_t$ for a given query camera $Q$:
\begin{equation}
    \widehat D_Q = G_{\psi}(S_t,Q).
    \label{eq:geometry_query}
\end{equation}
Thus, the same GeoState can be queried from different
viewpoints without re-encoding the scene.

\textbf{GeoState world modeling.}
With viewpoint variation removed from the predictive state, temporal modeling can focus on scene dynamics rather than camera-induced changes in the representation. Given four historical GeoStates $\{S_{t-3},S_{t-2},S_{t-1},S_t\}$, a temporal world model autoregressively predicts future states using a sliding four-state context. Each predicted state $\widehat S_{t+h}$ is then queried under a future camera $Q_{t+h}$ to obtain $\widehat D_{t+h}=G_{\psi}(\widehat S_{t+h},Q_{t+h})$. The query camera can be externally specified or provided by a separately trained camera prediction model, whose architecture and training details are given in Appendix~\ref{app:camera_predictor}.

\subsection{View-Decoupled Geometry State}
\label{sec:state}

Given synchronized multi-view images
$\mathcal{I}_t=\{I_t^v\}_{v\in\mathcal{V}_t}$, let
$\mathcal{I}^{\mathrm{in}}$ denote the observations jointly processed by
the frozen VGGT-$\Omega$~\citep{wang2026vggtomega}. To maintain a common
coordinate anchor across timestamps, $\mathcal{I}^{\mathrm{in}}$ may
include observations from multiple timestamps that share the same first
input view. After the joint VGGT-$\Omega$ forward pass, however, the
Scene Tokenizer uses only the outputs associated with the current
synchronized observations $\mathcal{I}_t$:
\begin{equation}
    \left\{
        \{F_t^{v,(\ell)}\}_{\ell\in\{4,11,17,23\}},
        X_t^v,C_t^v,\widetilde Q_t^v
    \right\}_{v\in\mathcal V_t}
    =
    \left.
    \mathcal G_\Omega(\mathcal I^{\mathrm{in}})
    \right|_t.
    \label{eq:vggt_outputs}
\end{equation}
Here, $F_t^{v,(\ell)}$ denotes intermediate patch features from the four
levels used by the VGGT-$\Omega$ dense prediction head, $X_t^v$ and
$C_t^v$ denote the pixel-aligned 3D point map and its confidence,
respectively, and
$\widetilde Q_t^v=(\widetilde K_t^v,\widetilde P_t^v)$ is the
GFM-estimated camera. Thus, cross-time observations are used by
VGGT-$\Omega$ to establish a shared coordinate frame, while each
GeoState is constructed only from the outputs corresponding to its own
timestamp. For the observed context states, all context observations are
processed in a single forward pass; future target states use growing
observation prefixes as described in Sec.~\ref{sec:training}. We omit
the time index $t$ below for notational simplicity.

\paragraph{Geometry-conditioned GFM tokens.}
For each patch $p$ in source view $v$, we project and fuse the GFM
features $\{F_{p,v}^{(\ell)}\}_{\ell\in\{4,11,17,23\}}$ with per-level
linear projections and an MLP to obtain
$e_{p,v}^{\mathrm{feat}}\in\mathbb{R}^{512}$. To facilitate cross-view feature aggregation and reduce scale inconsistency across GFM forwards,
following Surflo~\citep{guedon2026surflo}, we augment the GFM patch features with
explicit geometric cues derived from $X^v$, $C^v$, and
$\widetilde Q^v$. For each patch, we obtain its
3D location and confidence, source-camera pose, and viewing ray, and
encode them as $e_{p,v}^{\mathrm{point}}$,
$e_v^{\mathrm{pose}}$, and $e_{p,v}^{\mathrm{ray}}$, respectively,
after median-normalizing their explicit spatial scale. These embeddings form the explicit
geometric encoding
\begin{equation}
    g_{p,v}
    =
    [e_{p,v}^{\mathrm{point}},
     e_v^{\mathrm{pose}},
     e_{p,v}^{\mathrm{ray}}],
    \label{eq:geometric_encoding}
\end{equation}
which is fused with the GFM feature through
\begin{equation}
    m_{p,v}
    =
    \operatorname{LN}\!\left(
        e_{p,v}^{\mathrm{feat}}
        +
        \operatorname{MLP}\!\left(
            [e_{p,v}^{\mathrm{feat}},g_{p,v}]
        \right)
    \right).
    \label{eq:geometric_feature_fusion}
\end{equation}

\paragraph{Scene-state aggregation.}
To obtain a scene representation independent of the number of views, we collect the geometry-conditioned tokens from all source views into a variable-length memory
\begin{equation}
    \mathbf{M}_t
    =
    \operatorname{Concat}_{v,p} m_{p,v}
    \in\mathbb{R}^{N_t\times512}.
    \label{eq:geometry_memory}
\end{equation}
A Perceiver-style Tokenizer~\citep{jaegle2021perceiver} $E_{\phi}$ then aggregates this memory using
learned scene queries
$\mathbf{L}\in\mathbb{R}^{N_s\times d_s}$:
\begin{equation}
    S_t
    =
    E_{\phi}(\mathbf{M}_t;\mathbf{L})
    \in\mathbb{R}^{N_s\times d_s}.
    \label{eq:scene_tokenizer}
\end{equation}
We use $N_s=d_s=512$ and share the same learned scene queries across
timestamps. The resulting GeoState has a fixed size independent of the
number of source views. Detailed encoder
architecture is provided in Appendix~\ref{app:tokenizer_architecture}.

\subsection{Query-Conditioned Geometry Decoder}
\label{sec:decoder}

The query viewpoint is introduced only during geometry decoding.
Given a query camera $Q=(K,P)$, we construct a Pl\"ucker ray field~\citep{sitzmann2021lightfield}
from its intrinsics and camera-to-world pose. The rays are expressed
in the same coordinate frame and scale as the encoded scene
geometry, then Fourier-encoded~\citep{tancik2020fourier} and patchified into
query tokens $R_Q$.

We first process the scene state with a scene-only
self-attention conditioner, then concatenate it with $R_Q$.
A joint Transformer exchanges information among scene and
ray tokens to produce view-conditioned representations:
\begin{equation}
    Z_Q
    =
    \Psi_{\mathrm{query}}(S_t,R_Q),
    \label{eq:query_transformer}
\end{equation}
where $\Psi_{\mathrm{query}}$ denotes the query Transformer.
A feature decoder $\Phi_{\mathrm{dec}}$ maps $Z_Q$ to the four feature
levels expected by the frozen VGGT-$\Omega$ dense prediction head $H_{\mathrm{GFM}}$,
which produces the depth:
\begin{equation}
    \widehat D_Q^{\mathrm{norm}}
    =
    H_{\mathrm{GFM}}\!\left(
        \Phi_{\mathrm{dec}}(Z_Q)
    \right).
    \label{eq:decoder}
\end{equation}
The normalized prediction is converted back to metric depth using the
same context-derived scale convention used by the scene representation.
Reusing the frozen GFM head constrains the decoder to recover geometry-compatible features while avoiding relearning dense geometry prediction. Detailed model architectures
are provided in Appendix~\ref{app:decoder_architecture}.

\subsection{Scene-State World Model}
\label{sec:world_model}

The world model predicts future evolution directly in GeoState space.
By separating scene geometry from the query viewpoint, GeoState
allows camera-induced observation changes to be handled during decoding,
making scene persistence a natural prior for temporal prediction.
We therefore formulate future-state prediction as a residual update
to the latest state:
\begin{equation}
\widehat S_{t+1}
=
S_t+\Delta S_{t+1},
\qquad
\Delta S_{t+1}
=
F_{\theta}(S_{t-3:t}).
\label{eq:residual_prediction}
\end{equation}

We adopt the masked factorized Transformer architecture of
FR3D~\citep{morbitzer2026fr3d} for scene-state prediction.
A masked prediction state is appended to the four historical states,
followed by factorized temporal and scene attention. The predicted state
is produced through a residual head, which is zero-initialized so that
the model is exactly equivalent to latent Copy-Last at initialization,
and outputs $\Delta S_{t+1}$ in Eq.~\ref{eq:residual_prediction}.

Future GeoStates are generated autoregressively using a sliding
four-state context:
\begin{equation} \widehat S_{t+h} = \widetilde S_{t+h-1} + F_{\theta}\!\left( \widetilde S_{t+h-4:t+h-1} \right), \qquad h \geq 1, \label{eq:autoregressive_rollout} \end{equation}
where $\widetilde S_{\tau}=S_{\tau}$ for observed timestamps
$\tau\leq t$ and $\widetilde S_{\tau}=\widehat S_{\tau}$ otherwise.
Detailed world-model architecture is provided in Appendix~\ref{app:world_architecture}.

\subsection{Learning Objectives}
\label{sec:training}

\paragraph{Geometry-state learning.}
We train the scene tokenizer and decoder
for geometry reconstruction and novel-view querying, keeping
the GFM and its dense head frozen. The objective is
\begin{equation}
    \mathcal{L}_{\mathrm{rep}}
    =
    \mathcal{L}_{\log D}
    + \lambda_{\mathrm{grad}}\mathcal{L}_{\mathrm{grad}},
    \label{eq:representation_loss}
\end{equation}
where $\mathcal{L}_{\log D}$ measures mean absolute log-depth
error over valid pixels, and $\mathcal{L}_{\mathrm{grad}}$
supervises local depth gradients. The learned tokenizer and
decoder are frozen during world-model training.

\paragraph{World-model training.}
We train the world model in two stages: single-step initialization
followed by autoregressive rollout training.

\textit{Target-state construction.}
To construct future supervision in a consistent scene coordinate frame, target GeoStates are constructed by feeding growing observation prefixes
into the frozen VGGT-$\Omega$. For the target at $t+h$, the prefix
terminates at $t+h$, while all prefixes share the same first input view
to maintain a common coordinate anchor across time. Only the outputs
corresponding to the synchronized observations at $t+h$ are retained and
passed to the Scene Tokenizer to construct $S^*_{t+h}$.

Since VGGT-$\Omega$ predictions may vary in scale across forward passes,
we normalize the metric spatial quantities in $g_{p,v}$
(Eq.~\ref{eq:geometric_encoding}) using a calibration derived from the
observed context before fusing them with the GFM features. This gives
target GeoStates a consistent context-derived scale without target-frame
alignment. Further details are provided in Appendix~\ref{app:scale_protocol}.

\textit{Single-step initialization.}
We first train $F_{\theta}$ for one-step $4\!\rightarrow\!1$
prediction with joint state and geometric supervision:
\begin{equation}
    \mathcal{L}_{\mathrm{init}}
    =
    \lambda_{\mathrm{state}}\mathcal{L}_{\mathrm{state}}
    + \lambda_{\mathrm{depth}}
    \mathcal{L}_{\log D}(\widehat D_{t+1},D_{t+1}),
    \label{eq:init_loss}
\end{equation}
where $\mathcal{L}_{\mathrm{state}}$ combines Smooth L1 and
cosine losses between the predicted state $\widehat S_{t+1}$
and the encoded target $S^*_{t+1}$. The predicted state is also
decoded as
$\widehat D_{t+1}=G_{\psi}(\widehat S_{t+1},Q_{t+1})$
using the ground-truth future FRONT camera $Q_{t+1}$ and supervised
against LiDAR depth $D_{t+1}$. Thus, state supervision directly
guides prediction toward the encoded target, while depth supervision
constrains the geometry decoded from the predicted state.

\textit{Autoregressive rollout training.}
We then unroll the world model for $H=7$ future steps and supervise
each predicted state through its decoded geometry. At each step,
$\widehat D_{t+h}=G_{\psi}(\widehat S_{t+h},Q_{t+h})$ is decoded using
the ground-truth future FRONT camera $Q_{t+h}$:
\begin{equation}
    \mathcal{L}_{\mathrm{roll}}
    =
    \frac{1}{H}\sum_{h=1}^{H}
    \mathcal{L}_{\log D}
    (\widehat D_{t+h},D_{t+h}),
    \label{eq:rollout_loss}
\end{equation}
Unlike single-step initialization, we omit direct state matching during
rollout, allowing autoregressive states to be optimized for their decoded
geometry rather than exact agreement with GFM features.
We examine the role of state supervision in both training stages in
Section~\ref{sec:ablation}.

\section{Experiments}
\label{sec:experiments}

\subsection{Experimental Setup}
\label{sec:experimental_setup}
\paragraph{Datasets.}
We pretrain the Scene Tokenizer and ray-conditioned decoder on
approximately 10k multi-view scenes from
DL3DV~\citep{ling2024dl3dv}, TartanAir~\citep{wang2020tartanair},
RealEstate10K~\citep{zhou2018stereo}, CO3D~\citep{reizenstein2021co3d},
and WildRGB-D~\citep{xia2024rgbd}, using ground-truth depth where
available and VGGT-$\Omega$~\citep{wang2026vggtomega} pseudo-depth otherwise.
We then adapt both modules on a subset of the Waymo Open
Dataset~\citep{sun2020waymo} selected for static scene content. World Model training uses approximately
140k windows from 800 segments, each containing four context and seven
future states at $0.2$\,s intervals, with FRONT, FRONT\_LEFT, and
FRONT\_RIGHT context views. We evaluate on 100 Waymo windows held out from model selection,
113 eligible KITTI~\citep{geiger2012kitti} drives, and all 150 scenes
in the nuScenes~\citep{caesar2020nuscenes} validation split.
Further dataset selection and sampling details are provided in
Appendix~\ref{app:evaluation_protocol}.

\paragraph{Evaluation Metrics.}
We report frame-wise mean AbsRel ($\downarrow$) and $\delta_1$
($\uparrow$)~\citep{eigen2014depth} over valid pixels.
Our method flexibly adapts VGGT-$\Omega$'s default balanced resolution preprocessing to both training and evaluation, while baselines follow their
original resolution settings.
For scale handling, we follow the original evaluation procedure of each baseline: VGGT-World performs per-target-frame scale alignment, while FR3D performs per-target-frame scale-and-shift alignment. For DINO-Foresight, we fit its relative inverse-depth predictions to the LiDAR measurements of each target frame. In contrast, we calibrate metric scale once using LiDAR at the last observed FRONT frame and retain this scale throughout the entire forecasting rollout without target-frame alignment.

We additionally measure cross-query geometric consistency using symmetric
reprojection AbsRel. Given decoded depths $D_0,D_i$ and z-buffered
reprojection $\mathcal{R}_{a\rightarrow b}$, we compute
\begin{equation} E_{\mathrm{sym}}(Q_0,Q_i) = \frac{1}{2}\big[ \operatorname{AbsRel}\!\left(\mathcal{R}_{0\rightarrow i}(D_0),D_i\right) + \operatorname{AbsRel}\!\left(\mathcal{R}_{i\rightarrow0}(D_i),D_0\right) \big]. \label{eq:query_consistency} \end{equation}
Lower values indicate better geometric agreement.

\paragraph{Baselines.}
We compare with DINO-Foresight~\citep{karypidis2025dinoforesight},
which forecasts semantic image features, and geometry-based world
models FR3D~\citep{morbitzer2026fr3d} and VGGT-World~\citep{sun2026vggtworld}.
We train DINO-Foresight and VGGT-World on the same Waymo training data
as our method, but following their respective training procedures.
All FR3D results are taken directly from the original paper.

We also include three Copy-Last baselines.
Fixed View repeats the last observed VGGT-$\Omega$ depth at every future timestamp, while GT Camera and
Pred. Camera decode the last observed scene state from
ground-truth or predicted future cameras. These baselines allow us to assess the
effects of future camera motion and scene-state prediction.
\paragraph{Implementation Details.}
We pretrain the Scene Tokenizer and ray-conditioned decoder
on the multi-view pretraining data for 200k steps with an initial
learning rate of $10^{-5}$, and then finetune on the static
Waymo subset for 50k steps with an initial learning rate of
$5{\times}10^{-6}$. Both stages use a global effective batch
size of 36, with the VGGT-$\Omega$ encoder frozen. We sample one to five
source views per scene and supervise depth reconstruction at held-out
query views. For static Waymo sequences, source and query views may span
both cameras and timestamps. We then train the World Model in two stages:
20k steps of one-step prediction with an initial learning rate of
$10^{-5}$, followed by 35k steps of seven-step autoregressive rollout
training with an initial learning rate of $10^{-6}$. VGGT-$\Omega$, the Scene Tokenizer, and
the decoder remain frozen throughout World Model training, while gradients propagate through the
decoder to the predicted states.
Additional training details are provided in
Appendix~\ref{app:training_details}.

\subsection{Long-Horizon Depth Forecasting on Waymo}
\label{sec:waymo_forecasting}

\begin{table}[t]
\centering
\caption{\textbf{Long-horizon depth forecasting on Waymo.}
Bold and underlined values indicate the best and second-best results among methods using predicted future cameras.}
\label{tab:main_waymo}

\begingroup
\footnotesize
\setlength{\tabcolsep}{2.3pt}
\renewcommand{\arraystretch}{1.02}
\setlength{\aboverulesep}{1.5pt}
\setlength{\belowrulesep}{1.5pt}
\setlength{\cmidrulesep}{1pt}

\begin{adjustbox}{max width=\linewidth}
\begin{tabular}{@{}l c c *{4}{cc}@{}}
\toprule
\textbf{Method}
& \textbf{Camera}
& \textbf{Per-frame}
& \multicolumn{2}{c}{$t{+}0.2$ s}
& \multicolumn{2}{c}{$t{+}0.6$ s}
& \multicolumn{2}{c}{$t{+}1.0$ s}
& \multicolumn{2}{c}{$t{+}2.0$ s} \\
\cmidrule(lr){4-5}
\cmidrule(lr){6-7}
\cmidrule(lr){8-9}
\cmidrule(l){10-11}
& & \textbf{alignment}
& AbsRel $\downarrow$ & $\delta_1 \uparrow$
& AbsRel $\downarrow$ & $\delta_1 \uparrow$
& AbsRel $\downarrow$ & $\delta_1 \uparrow$
& AbsRel $\downarrow$ & $\delta_1 \uparrow$ \\
\midrule

Copy-Last & Fixed View & No
& 0.094 & 0.906
& 0.147 & 0.840
& 0.184 & 0.797
& 0.248 & 0.734 \\

\midrule

DINO-Foresight & Pred. & Yes
& 0.119 & 0.841
& 0.138 & 0.825
& 0.151 & 0.798
& 0.243 & 0.663 \\

VGGT-World & Pred. & Yes
& 0.073 & 0.931
& 0.112 & 0.892
& 0.143 & 0.826
& 0.213 & 0.731 \\

FR3D & Pred. & Yes
& 0.110 & 0.879
& 0.121 & 0.858
& 0.131 & 0.840
& 0.152 & 0.800 \\

Copy-Last & Pred. & No
& \underline{0.071} & \underline{0.933}
& \underline{0.088} & \underline{0.905}
& \underline{0.100} & \underline{0.882}
& \underline{0.132} & \underline{0.827} \\

\rowcolor{oursblue}
GeoState & Pred. & No
& \textbf{0.065} & \textbf{0.939}
& \textbf{0.079} & \textbf{0.921}
& \textbf{0.092} & \textbf{0.902}
& \textbf{0.127} & \textbf{0.848} \\

\midrule

Copy-Last & GT & No
& 0.068 & 0.934
& 0.082 & 0.913
& 0.092 & 0.894
& 0.108 & 0.865 \\

\rowcolor{oursblue}
GeoState & GT & No
& 0.062 & 0.940
& 0.071 & 0.929
& 0.081 & 0.915
& 0.099 & 0.888 \\

\bottomrule
\end{tabular}
\end{adjustbox}
\endgroup
\end{table}
\begin{figure}[t]
\centering
\includegraphics[width=\linewidth]{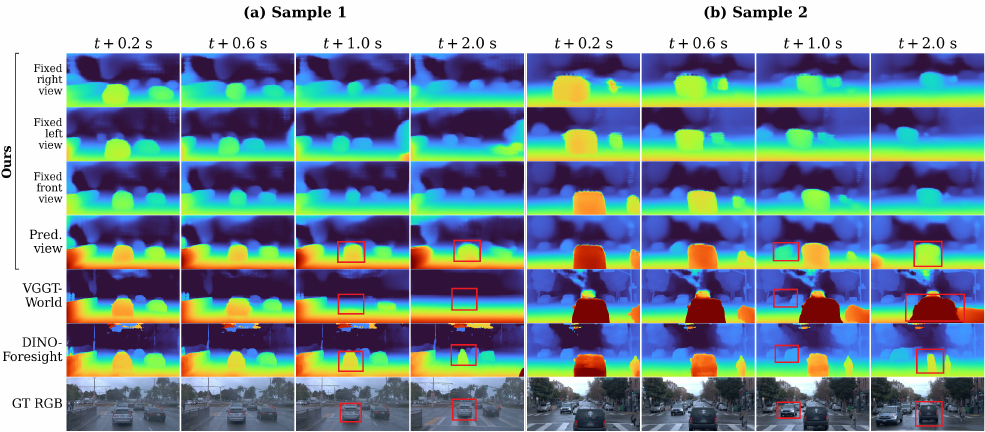}
\caption{\textbf{Qualitative multi-view depth forecasting.}
Red boxes highlight prediction differences.}
\label{fig:multiview_forecasting}
\end{figure}

Table~\ref{tab:main_waymo} compares long-horizon depth forecasting performance on Waymo. With predicted cameras, our model consistently outperforms all compared methods on both depth metrics at every reported horizon. The advantage over VGGT-World increases with the forecasting horizon, reaching a $40.4\%$ reduction in AbsRel at $2.0$\,s. These improvements are achieved while maintaining a single context-derived scale without per-frame alignment and persist beyond the $1.4$\,s rollout horizon used during training. These results demonstrate that GeoState supports effective long-horizon autoregressive geometry forecasting, including at horizons beyond those seen during training. Figure~\ref{fig:multiview_forecasting} visualizes depth forecasts across future horizons and query views, alongside comparisons with prior methods.

The Copy-Last variants further disentangle the effects of camera conditioning and scene-state evolution. Re-querying the last observed state with future cameras substantially outperforms simply repeating the last observed depth map, demonstrating that viewpoint-induced changes can be accounted for by querying a fixed scene state from future viewpoints without evolving the state itself. Our predicted states consistently outperform their camera-matched Copy-Last counterparts at every horizon under both GT and predicted cameras, showing that the World Model captures future scene evolution beyond what can be explained by camera re-querying alone.
Motion-stratified evaluation further shows that these improvements are more pronounced in Dynamic than in Static windows under both GT and predicted cameras, supporting the effectiveness of GeoState in modeling scene dynamics. Detailed
results are provided in Appendix~\ref{app:Motion-Stratified Forecasting Results}.

\subsection{Cross-Query Geometric Consistency}
\label{sec:cross_query_consistency}

We evaluate whether a predicted scene state remains geometrically consistent
when queried from different viewpoints. At each future timestamp, we decode
the same state from the ground-truth FRONT camera $Q_0$ and perturbed virtual
cameras $Q_i$, jointly varying translation and rotation. We consider Small
($0$--$0.5$\,m, $0$--$2^\circ$), Medium
($0.5$--$1.0$\,m, $2$--$5^\circ$), and Large
($1.0$--$1.5$\,m, $5$--$10^\circ$) perturbations.
Consistency is measured by symmetric reprojection AbsRel
$E_{\mathrm{sym}}$ in Eq.~\ref{eq:query_consistency} over valid overlap.
We compare World Model-predicted states (World) with target-time states
encoded from RGB observations (Enc.). Further protocol details are provided in
Appendix~\ref{app:query_consistency}.

\begin{figure}[t]
  \centering
  \includegraphics[width=\linewidth]{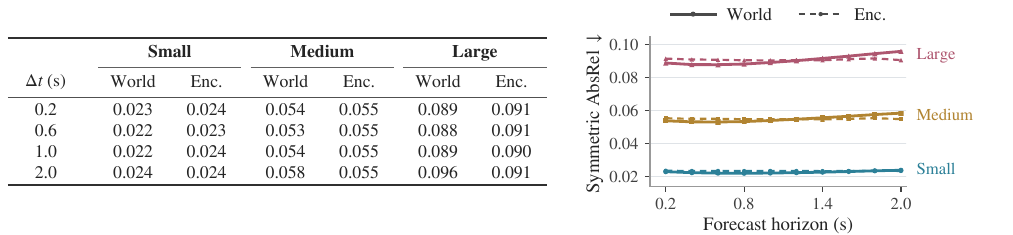}
  \caption{\textbf{Cross-query geometric consistency on Waymo.}
We evaluate symmetric reprojection AbsRel ($\downarrow$) across perturbation levels.}
  \label{fig:query_consistency_horizons}
\end{figure}
\begin{table}[t]
\centering
\caption{\textbf{Zero-shot depth forecasting on KITTI and nuScenes.}
Bold and underlined values indicate the best and second-best results
among methods using predicted cameras.}
\label{tab:zero_shot}

\begingroup
\footnotesize
\setlength{\tabcolsep}{2.2pt}
\renewcommand{\arraystretch}{1.0}
\setlength{\aboverulesep}{1.5pt}
\setlength{\belowrulesep}{1.5pt}
\setlength{\cmidrulesep}{1pt}

\begin{adjustbox}{max width=\linewidth}
\begin{tabular}{@{}l c *{3}{cc} @{\hspace{8pt}} *{3}{cc}@{}}
\toprule
\multicolumn{2}{@{}l}{}
& \multicolumn{6}{c}{\textbf{KITTI}}
& \multicolumn{6}{c}{\textbf{nuScenes}} \\
\cmidrule(lr){3-8}
\cmidrule(l){9-14}

\textbf{Method}
& \textbf{Camera}
& \multicolumn{2}{c}{$t{+}0.6$ s}
& \multicolumn{2}{c}{$t{+}1.0$ s}
& \multicolumn{2}{c}{$t{+}2.0$ s}
& \multicolumn{2}{c}{$t{+}0.75$ s}
& \multicolumn{2}{c}{$t{+}1.25$ s}
& \multicolumn{2}{c}{$t{+}2.5$ s} \\
\cmidrule(lr){3-4}
\cmidrule(lr){5-6}
\cmidrule(lr){7-8}
\cmidrule(lr){9-10}
\cmidrule(lr){11-12}
\cmidrule(l){13-14}

&
& AbsRel$\downarrow$ & $\delta_1\uparrow$
& AbsRel$\downarrow$ & $\delta_1\uparrow$
& AbsRel$\downarrow$ & $\delta_1\uparrow$
& AbsRel$\downarrow$ & $\delta_1\uparrow$
& AbsRel$\downarrow$ & $\delta_1\uparrow$
& AbsRel$\downarrow$ & $\delta_1\uparrow$ \\

\midrule

DINO-Foresight & Pred.
& 0.131 & 0.849
& 0.161 & 0.793
& 0.195 & 0.732
& 0.231 & 0.691
& 0.253 & 0.665
& 0.291 & 0.602 \\

VGGT-World & Pred.
& {0.109} & {0.886}
& {0.127} & {0.862}
& {0.198} & {0.732}
& {0.228} & {0.715}
& {0.241} & {0.693}
& {0.282} & {0.637} \\

FR3D & Pred.
& 0.116 & 0.868
& 0.132 & 0.835
& 0.178 & 0.758
& 0.182 & 0.740
& 0.197 & 0.710
& 0.229 & \underline{0.660} \\

Copy-Last & Pred.
& \underline{0.097} & \underline{0.913}
& \underline{0.113} & \underline{0.878}
& \underline{0.160} & \underline{0.802}
& \underline{0.142} & \underline{0.804}
& \underline{0.177} & \underline{0.751}
& \underline{0.221} & 0.646 \\

\rowcolor{oursblue}
{GeoState} & Pred.
& \textbf{0.095} & \textbf{0.920}
& \textbf{0.112} & \textbf{0.883}
& \textbf{0.155} & \textbf{0.821}
& \textbf{0.136} & \textbf{0.820}
& \textbf{0.172} & \textbf{0.764}
& \textbf{0.213} & \textbf{0.666} \\

\midrule

Copy-Last & GT
& 0.088 & 0.932
& 0.095 & 0.915
& 0.114 & 0.882
& 0.131 & 0.825
& 0.164 & 0.786
& 0.195 & 0.697 \\

\rowcolor{oursblue}
GeoState & GT
& 0.086 & 0.939
& 0.093 & 0.918
& 0.108 & 0.895
& 0.125 & 0.840
& 0.159 & 0.797
& 0.190 & 0.707 \\

\bottomrule
\end{tabular}
\end{adjustbox}

\endgroup
\end{table}

Figure~\ref{fig:query_consistency_horizons} shows that world-model-predicted states closely match observation-encoded states across all perturbation levels at every forecast horizon.
This indicates that although geometric supervision is provided only through
depth, the model learns a scene state that captures the underlying geometric
structure of the world and supports consistent querying from unseen cameras.

\subsection{Zero-Shot Generalization}
\label{sec:zero_shot}

Table~\ref{tab:zero_shot} evaluates zero-shot depth forecasting on KITTI
and nuScenes. Without target-dataset adaptation, our model with predicted
cameras outperforms DINO-Foresight, VGGT-World, and the reported FR3D
results on both metrics across all reported horizons on both datasets.
These results support that GeoState captures transferable geometric
structure. Moreover, our
predicted states consistently outperform camera-matched Copy-Last under
both ground-truth and predicted cameras, showing that the learned scene
dynamics also generalize beyond the training domain.
\begin{table}[t]
\centering
\caption{\textbf{Geometric reconstruction on Waymo.}}
\vspace{3pt}
\label{tab:ablation_oracle}

\begingroup
\footnotesize
\setlength{\tabcolsep}{3.0pt}
\renewcommand{\arraystretch}{0.76}

\setlength{\aboverulesep}{0.8pt}
\setlength{\belowrulesep}{0.8pt}
\setlength{\cmidrulesep}{0pt}

\begin{tabular*}{\linewidth}
{@{}l@{\extracolsep{\fill}}*{4}{cc}@{}}
\toprule
\textbf{Method}
& \multicolumn{2}{c}{$t{+}0.2$ s}
& \multicolumn{2}{c}{$t{+}0.6$ s}
& \multicolumn{2}{c}{$t{+}1.0$ s}
& \multicolumn{2}{c}{$t{+}2.0$ s} \\

\cmidrule(lr){2-3}
\cmidrule(lr){4-5}
\cmidrule(lr){6-7}
\cmidrule(l){8-9}

& AbsRel $\downarrow$ & $\delta_1 \uparrow$
& AbsRel $\downarrow$ & $\delta_1 \uparrow$
& AbsRel $\downarrow$ & $\delta_1 \uparrow$
& AbsRel $\downarrow$ & $\delta_1 \uparrow$ \\
\midrule

VGGT-$\Omega$
& 0.074 & 0.934
& 0.079 & 0.932
& 0.076 & 0.932
& 0.078 & 0.930 \\

GeoState Reconstruction
& 0.068 & 0.934
& 0.073 & 0.932
& 0.072 & 0.932
& 0.072 & 0.929 \\

\bottomrule
\end{tabular*}

\endgroup
\end{table}

\begin{table}[t]
\centering
\begingroup
\setlength{\tabcolsep}{0.8pt}
\renewcommand{\arraystretch}{1.08}
\setlength{\abovecaptionskip}{2pt}
\setlength{\belowcaptionskip}{4pt}

\begin{minipage}[t]{0.49\linewidth}
\vspace{0pt}
\centering
\caption{\textbf{Representation ablation.}}
\label{tab:ablation_representation}

\scriptsize
\begin{tabular*}{\linewidth}
{@{}@{\extracolsep{\fill}}ll*{3}{cc}@{}}
\toprule
\textbf{Mode} & \textbf{State}
& \multicolumn{2}{c}{$t{+}0.6$\,s}
& \multicolumn{2}{c}{$t{+}1.0$\,s}
& \multicolumn{2}{c}{$t{+}2.0$\,s} \\
\cmidrule(lr){3-4}
\cmidrule(lr){5-6}
\cmidrule(l){7-8}
& &
AbsRel$\downarrow$ & $\delta_1\uparrow$ &
AbsRel$\downarrow$ & $\delta_1\uparrow$ &
AbsRel$\downarrow$ & $\delta_1\uparrow$ \\
\midrule

Rec. & Compressor
& 0.075 & 0.930
& 0.073 & 0.930
& 0.074 & 0.928 \\
& Ours
& 0.073 & 0.932
& 0.072 & 0.932
& 0.072 & 0.929 \\

\addlinespace[2pt]

Pred. & Compressor
& 0.092 & 0.896
& 0.111 & 0.863
& 0.144 & 0.804 \\
& Ours
& \textbf{0.081} & \textbf{0.919}
& \textbf{0.095} & \textbf{0.896}
& \textbf{0.133} & \textbf{0.834} \\

\bottomrule
\end{tabular*}
\end{minipage}\hfill%
\begin{minipage}[t]{0.49\linewidth}
\vspace{0pt}
\centering
\caption{\textbf{State supervision ablation.}}
\label{tab:ablation_supervision}

\scriptsize
\begin{tabular*}{\linewidth}
{@{}@{\extracolsep{\fill}}cc*{3}{cc}@{}}
\toprule
\multicolumn{1}{@{}c}{\textbf{One-step}}
& \multicolumn{1}{c}{\textbf{Rollout}}
& \multicolumn{2}{c}{$t{+}0.6$\,s}
& \multicolumn{2}{c}{$t{+}1.0$\,s}
& \multicolumn{2}{c}{$t{+}2.0$\,s} \\
\cmidrule(lr){3-4}
\cmidrule(lr){5-6}
\cmidrule(l){7-8}
& &
AbsRel$\downarrow$ & $\delta_1\uparrow$ &
AbsRel$\downarrow$ & $\delta_1\uparrow$ &
AbsRel$\downarrow$ & $\delta_1\uparrow$ \\
\midrule
$\times$ & $\checkmark$
& 0.077 & 0.919
& 0.086 & 0.902
& 0.108 & 0.878 \\
$\times$ & $\times$
& 0.075 & 0.922
& 0.084 & 0.909
& 0.104 & 0.879 \\
\addlinespace[2pt]
$\checkmark$ & $\checkmark$
& 0.071 & 0.928
& 0.082 & 0.913
& 0.107 & 0.879 \\
$\checkmark$ & $\times$
& \textbf{0.071} & \textbf{0.929}
& \textbf{0.081} & \textbf{0.915}
& \textbf{0.099} & \textbf{0.888} \\
\bottomrule
\end{tabular*}
\end{minipage}

\par\vspace{3pt}
\endgroup
\end{table}
\subsection{Ablation Studies}
\label{sec:ablation}

\paragraph{Geometric Reconstruction Quality.}
We first examine whether compressing GFM features into GeoState preserves
their geometric information. Table~\ref{tab:ablation_oracle} compares
direct VGGT-$\Omega$ depth predictions with depth reconstructed through
our Scene Tokenizer and Geometry Decoder using ground-truth query cameras.
Both use target-time RGB observations and are evaluated against
ground-truth LiDAR depth. Their comparable accuracy shows that the
fixed-length GeoState preserves the geometric information required for
accurate reconstruction, providing a compact and geometrically faithful
representation for subsequent world modeling.

\paragraph{Effect of the View-Decoupled Scene State.}
We compare GeoState with a feature-compression baseline that directly
compresses VGGT-$\Omega$ features into a $512\times512$ latent state. For a fair comparison with the feature-compression baseline, only the target FRONT GFM output
block is passed to the Scene Tokenizer. Both representations use the same world model,
training data, forecasting objectives, optimization, and evaluation protocol;
implementation details are provided in Appendix~\ref{app:compressor}. As shown in Table~\ref{tab:ablation_representation}, the two representations achieve comparable reconstruction quality, but GeoState is
substantially easier to predict, improving AbsRel and $\delta_1$ at
every horizon.

Because the two representations achieve comparable reconstruction quality, the forecasting
gap is unlikely to arise from reconstruction
accuracy. Instead, the results suggest that GeoState provides a more
predictable state space for temporal modeling. By separating scene content
from viewpoint-dependent observation, GeoState reduces camera-induced
variation in the state dynamics, allowing the World Model to focus more
directly on the underlying evolution of scene geometry.

\paragraph{Effect of State Supervision.}
We ablate state supervision during one-step initialization and
autoregressive rollout while retaining decoded-depth supervision in all
settings. Table~\ref{tab:ablation_supervision} shows that one-step state
supervision consistently improves forecasting, providing a useful target
for initializing local state transitions. In contrast, extending latent
matching throughout the rollout degrades
long-horizon performance. This may be attributed to context-dependent variations
in GFM features that do not necessarily correspond to
changes in scene geometry, as further analyzed in
Appendix~\ref{app:gfm_variation}.
These findings support using state matching to initialize local
transitions while supervising autoregressive predictions through
decoded geometry rather than enforcing a unique latent trajectory.

\section{Conclusion}

We presented GeoState, a view-decoupled geometry state learned from GFM
features for predictive 3D world modeling. By separating scene evolution
from camera-dependent observation, GeoState provides a scene representation that can be predicted more effectively over time and queried from different viewpoints.
Future work may extend this framework to embodied settings,
where predictive scene states could support perception,
planning, and interaction in dynamic environments.

\bibliography{references/references}
\bibliographystyle{iclr2027_conference}

\clearpage
\appendix

\section{Appendix}
\label{app}

\subsection{Architecture Details}
\label{app:architecture}

This section provides the implementation details of the Scene Tokenizer, geometry decoder, and world model used in the main experiments. Table~\ref{tab:app_architecture} summarizes their principal architectural configurations and parameter counts.

\subsubsection{Geometry-Conditioned Memory and Scene Tokenizer}
\label{app:tokenizer_architecture}

Each selected GFM feature level is independently normalized and projected
from 2,048 to 512 channels. The four projected features are concatenated
and fused by an MLP with dimensions $2048\rightarrow2048\rightarrow512$.
For each spatial patch, confidence-weighted pooling of the dense point map
provides a representative 3D point. Its coordinates, normalized by the
scene scale, are Fourier-encoded with ten bands and concatenated with
$\log(1+c)$, where $c$ is the pooled confidence, before projection to a
512-dimensional geometry embedding.

Source-camera conditioning uses a six-dimensional rotation representation
formed from the first two rotation-matrix columns and a four-band Fourier
encoding of normalized translation. A camera MLP produces a
512-dimensional embedding shared by all patches in that view.
The source-ray branch encodes six-dimensional Pl\"ucker coordinates with
four Fourier bands and applies a convolution with kernel size and stride
16, followed by LayerNorm, to produce one 512-dimensional embedding per patch.

The feature, point, camera, and ray embeddings are concatenated. A
$2048\rightarrow1024\rightarrow512$ fusion MLP predicts a correction to
the feature embedding; their sum is normalized to form the patch memory.
A bank of 512 learned queries of width 512 is updated by eight Perceiver
layers, each comprising latent-to-memory cross-attention and latent
self-attention. Each attention block uses eight heads, pre-normalization,
residual connections, and a feed-forward network with expansion ratio four.
A final LayerNorm produces $S\in\mathbb{R}^{512\times512}$.

\subsubsection{Query-Conditioned Geometry Decoder}
\label{app:decoder_architecture}

For a camera-to-world pose $P=(R,\mathbf{o})$ and intrinsic matrix $K$,
a pixel $\mathbf{u}=(u,v,1)^\top$ defines the unit direction
\begin{equation}
\mathbf{d}(\mathbf{u})
=
\frac{RK^{-1}\mathbf{u}}{\|RK^{-1}\mathbf{u}\|_2}.
\end{equation}
In normalized scene coordinates, the ray representation is
$[\mathbf{d},\bar{\mathbf{o}}\times\mathbf{d}]$, where
$\bar{\mathbf{o}}$ is the normalized camera center.
Query rays are Fourier-encoded with six bands and patchified by a
stride-16 convolution. At $416\times624$ resolution, this produces
$26\times39=1{,}014$ query tokens, which are concatenated with the
512 scene tokens to form a sequence of 1,526 tokens.

The scene tokens first pass through LayerNorm, a linear projection,
and two scene-only self-attention blocks. The resulting scene tokens
and query-ray tokens are then processed jointly by sixteen
self-attention Transformer blocks, allowing each query to retrieve
view-dependent geometry from the shared scene state. Query-token
representations are retained after blocks 4, 8, 12, and 16, providing
multi-layer features at four stages of the decoder.

These multi-layer features are subsequently processed by a
\emph{DPT Feature Adapter}, whose role is to transform the decoder
representations into the feature interface expected by the frozen
VGGT-$\Omega$ DPT head. For each level, the final query representation
is fused with the corresponding intermediate representation through
a learned sigmoid gate and a level embedding. Two additional
Perceiver layers allow the query tokens to attend to the final scene
tokens, and an independent MLP maps each level as
$512\rightarrow1024\rightarrow1024\rightarrow2048$.

The resulting four feature maps are passed directly to the frozen
VGGT-$\Omega$ DPT head to predict camera $z$-depth. The decoder receives only the scene
state, query camera, and output dimensions, such that all scene
information must pass through the fixed-size state bottleneck.
During world-model training, both the Geometry Decoder and DPT head
remain fixed, while gradients are propagated through them with
respect to the input scene state.

\subsubsection{Residual World Model}
\label{app:world_architecture}
\begin{table}[t]
\centering

\setlength{\belowcaptionskip}{6pt}

\caption{Detailed architecture of the model.}
\label{tab:app_architecture}

\small
\setlength{\aboverulesep}{2.5pt}
\setlength{\belowrulesep}{2.5pt}
\renewcommand{\arraystretch}{1.05}
\begin{tabular}{@{}lp{0.62\linewidth}@{}}
\toprule
Component & Configuration \\
\midrule
Scene Tokenizer & 512 queries of width 512; eight Perceiver layers;
eight attention heads; MLP expansion ratio 4; 70.46M parameters \\
Geometry decoder & Width 512; two scene-conditioning blocks;
16 joint scene--ray blocks; four output branches with two
Perceiver layers each; 135.55M parameters \\
World Model & Four-state context; 12 factorized blocks; width 1,152;
eight heads; MLP expansion ratio 4; dropout 0.1;
256.55M parameters \\

\bottomrule
\end{tabular}
\end{table}

The four observed scene states form an input tensor of shape
$4\times512\times512$, corresponding to four time steps, 512 spatial
slots per state, and 512 feature channels per slot. Each token is first
linearly projected from 512 to a model width of 1,152. A learned mask
token is then replicated over 512 slots to construct a fifth state,
which serves as the prediction position. To encode token identity,
we use learned temporal and slot embeddings of 576 dimensions each.
Their concatenation forms a 1,152-dimensional positional embedding,
which is added to the projected tokens.

Each Transformer block factorizes attention over time and space.
Temporal attention is first applied across the five time positions
independently for each of the 512 slots. Spatial attention then operates
across the 512 slots independently within each time position. The two
attention operations are followed by a
$1152\rightarrow4608\rightarrow1152$ feed-forward network with GELU.
All sublayers use pre-normalization, residual connections, and dropout.
After the final block, the tokens at the fifth, masked prediction
position are mapped back to 512 channels by a bias-free linear residual
head. The head is zero-initialized, such that the model initially
predicts a zero residual and therefore exactly recovers the Copy-Last
prediction.

\subsection{Camera Predictor}
\label{app:camera_predictor}

The camera predictor forecasts future FRONT-camera poses from the observed ego-motion and the current FRONT image. The predicted trajectory is used as the future query-camera sequence for both GeoState (Pred. Camera) and Copy-Last (Pred. Camera).

\subsubsection{Architecture}
\label{app:camera_architecture}

The camera predictor contains 8.99M parameters and follows a
motion-prior-plus-visual-refinement design. Given four historical metric
FRONT poses at $t-0.6$, $t-0.4$, $t-0.2$, and $t$ seconds, we first compute
the three consecutive relative transformations and map them to
six-dimensional motion vectors using the $\mathrm{SE}(3)$ logarithm.
The motion vectors and their first differences are standardized, projected
to width 256, and encoded by a four-layer Transformer with eight attention
heads, an MLP expansion ratio of four, and dropout 0.1. Future poses are initialized from a constant-body-twist extrapolation of
the latest observed motion. Learned horizon queries attend to the encoded
motion features through three decoding layers and predict residual
corrections to this extrapolated trajectory. Future time is represented
using linear and quadratic terms together with eight Fourier bands.

To incorporate scene-dependent cues, the predictor additionally conditions
on the current FRONT image. Frozen VGGT-$\Omega$ spatial features are
extracted from a separate forward pass and pooled to a $6\times8$ grid,
yielding 48 tokens of width 2,048. After projection to width 256 and the
addition of learned spatial embeddings, two cross-attention layers compress
these features into eight visual tokens. The visual tokens are appended to
the encoded motion memory, and two additional horizon-query decoding layers,
with eight attention heads and an MLP expansion ratio of two, predict a
six-dimensional visual correction $\Delta\xi_h$ for each requested horizon.

For each requested horizon $h$, the final relative pose is
\begin{equation}
  \widehat P_{t+h}
  = \widehat P^{\mathrm{motion}}_{t+h}
    \operatorname{Exp}_{\mathrm{SE}(3)}(\Delta\xi_h).
\end{equation}
The visual correction is bounded by a scaled $\tanh$ output, with a limit
of 2.0 in normalized residual coordinates and horizon-dependent motion
scales. Its final projection is initialized to zero, so the complete model
initially reduces to the motion-only prediction and learns visual corrections
during training. At inference, we use direct multi-horizon camera prediction for Waymo and KITTI, while nuScenes adopts autoregressive prediction at 0.25 s intervals, recursively updating the pose history with predicted poses.

\subsubsection{Training}
\label{app:camera_training}

We train the camera predictor in three stages. First, we optimize the motion
backbone using metric FRONT-camera trajectories from Waymo. Each training
sample contains four observed poses at $t{-}0.6$, $t{-}0.4$, $t{-}0.2$, and
$t$, together with ten future poses at $0.2\,\mathrm{s}$ intervals. The objective
combines absolute pose error, adjacent relative-pose error, and ray-endpoint
error with weights 1.0, 0.5, and 0.2, respectively. Rotation errors are
converted to metric displacement using a $10\,\mathrm{m}$ lever arm.
We first train for 12,000 steps with direct ten-horizon supervision using
AdamW, batch size 1,024, weight decay 0.05, and a cosine learning-rate
schedule from $10^{-4}$ to $10^{-6}$ after 300 warm-up steps. We then
continue for another 12,000 steps with closed-loop training, recursively
applying the predictor over four consecutive rollout chunks. The
closed-loop loss is weighted by 0.5. This phase uses batch size 384, a
100-step warm-up, and a cosine learning-rate schedule from
$5\times10^{-5}$ to $5\times10^{-7}$. Gradients are clipped to norm 1.0
throughout motion training.

Second, we train the visual adapter while keeping the motion backbone
frozen. We use the same pose-based supervision, with the relative-pose and
ray-endpoint terms both weighted by 0.5 and a $20\,\mathrm{m}$ rotation
lever arm. Turning and speed-changing windows are assigned sample weights
of 1.5 and 0.5, respectively. Training uses 123,789 windows for 20 epochs
with AdamW, batch size 256, weight decay 0.02, and a cosine learning-rate
schedule from $2\times10^{-4}$ to $2\times10^{-6}$. Gradients are clipped
to norm 1.0.

Finally, we adapt the visual adapter to the downstream depth-query task
through the frozen geometry decoder. The last observed GeoState $S_t$ is
held fixed across future horizons and decoded using the predicted future
camera poses. The decoded depth is supervised against future Waymo LiDAR. To account
for the horizon-dependent difficulty of querying a fixed scene state, we
normalize the AbsRel at each horizon by the corresponding GT-camera
Copy-Last AbsRel estimated over the training set. This depth objective is combined
with the trajectory loss and a parameter-anchor regularizer, weighted by
0.05 and 0.02, respectively. Only the visual adapter is updated in this
stage. We train for 1,200 steps using AdamW with learning rate
$5\times10^{-6}$, weight decay 0.01, and gradient clipping at 1.0.

\subsubsection{Evaluation Results}
\label{app:camera_results}

We evaluate on the same Waymo validation windows
as depth forecasting. Predicted and GT camera-to-world poses share the
metric frame of the last observed FRONT camera, with no post-hoc
trajectory alignment. At each reported horizon, errors are pooled over
all 100 sequences and all future steps up to that horizon before computing
RMSE. ATE measures camera-center error. RTE and RRE measure the translation
magnitude and rotation angle of
$(T_{k-1}^{-1}T_k)^{-1}
(\widehat T_{k-1}^{-1}\widehat T_k)$
over adjacent $0.2$\,s transitions, including the first transition from
the observed anchor. Table~\ref{tab:app_camera_trajectory_metrics}
reports the results.

\begin{table}[t]
\centering
\caption{Camera prediction results on Waymo.}
\label{tab:app_camera_trajectory_metrics}
\small
\setlength{\tabcolsep}{10pt}
\begin{tabular}{@{}rccc@{}}
\toprule
Horizon (s) & ATE (m) $\downarrow$ & RTE (m) $\downarrow$
& RRE ($^\circ$) $\downarrow$ \\
\midrule
0.2 & 0.0125 & 0.0125 & 0.1894 \\
0.6 & 0.0379 & 0.0246 & 0.2462 \\
1.0 & 0.0840 & 0.0361 & 0.2783 \\
2.0 & 0.3117 & 0.0785 & 0.3755 \\
\bottomrule
\end{tabular}
\end{table}

\subsection{Coordinate and Scale Conventions}
\label{app:scale_protocol}

GFM reconstruction is defined up to an unknown scale. For each temporal
window, we therefore calibrate its scale using the latest observed FRONT
view, which serves as the anchor. Let $D^\Omega_0$ and
$D^{\mathrm{LiDAR}}_0$ denote the raw GFM and LiDAR depths at the anchor,
respectively, and let $\Omega_0$ denote the set of valid pixels with matched
GFM and LiDAR depths. We estimate
\begin{equation}
a_0
=
\exp\!\left[
\operatorname{median}_{p\in\Omega_0}
\left(
\log D^\Omega_0(p)
-
\log D^{\mathrm{LiDAR}}_0(p)
\right)
\right],
\label{eq:app_anchor_scale}
\end{equation}
where $a_0$ converts metric distances to GFM units. Matched pixels are
required to have finite and positive GFM depth, LiDAR depth, and GFM
confidence, and LiDAR depth is restricted to 80\,m. Calibration is performed
only when at least 128 valid matches are available.

We additionally define a scene-normalization depth $s_0$ from the observed
views as a confidence-weighted trimmed median of valid GFM depths. The lowest
and highest 2\% of cumulative confidence weight are discarded before
computing the weighted median. The physical length represented by one
normalized scene unit is therefore
\begin{equation}
L_0
=
\frac{s_0}{a_0}
\quad \text{meters}.
\label{eq:app_physical_unit}
\end{equation}
Point coordinates, camera translations, and ray origins are divided by
$s_0$ in the reference GFM frame, while rotations and unit ray directions
remain unchanged. Pl\"ucker moments are constructed using the normalized
ray origins.

All four observed states within a temporal window are extracted from the
same 12-image GFM forward pass and therefore share the same $a_0$, $s_0$,
and $L_0$. A normalized decoder prediction is converted to metric depth as
\begin{equation}
\widehat D^{\mathrm{metric}}_Q
=
L_0 \widehat D^{\mathrm{norm}}_Q.
\label{eq:app_metric_depth}
\end{equation}
This conversion is fixed by the observed context and retained throughout the
entire forecasting rollout; no future target frame is used to re-estimate
the scale.

When geometry is obtained from a separate GFM forward pass, its raw scale
may differ from that of the observed context. Let $a_k$ denote the metric
calibration factor of such a pass. Setting
\begin{equation}
s_k = a_k L_0
\end{equation}
expresses its geometry in the same normalized physical unit, since
\begin{equation}
\frac{\mathbf{x}_{\mathrm{GFM},k}}{s_k}
\approx
\frac{\mathbf{x}_{\mathrm{metric}}}{L_0}.
\end{equation}
This normalization applies only to explicit geometric quantities (Eq.~\ref{eq:geometric_encoding}); the
intermediate GFM feature vectors are not rescaled.

\subsection{Training Details}
\label{app:training_details}
In this section, we provide the sampling procedures, loss implementations, and optimization
settings that supplement Sections~\ref{sec:training}
and~\ref{sec:experimental_setup}.
\subsubsection{Waymo Data Construction}
\label{app:Waymo Data Construction}
We use a binary motion partition for Waymo. Motion labels are derived from
the annotated 3D tracks of vehicles, pedestrians, and cyclists over each
eleven-state window (four observed and seven future states at 0.2\,s
intervals). A box observation must contain at least three LiDAR
points and a track must be observed in at least two states.
We mark a track as moving when its median speed exceeds $0.75$\,m/s for a
vehicle, $0.25$\,m/s for a pedestrian, or $0.50$\,m/s for a cyclist. A
window is Dynamic if it contains at least one such visible moving track and
Static otherwise. This partition yields 114,735 Dynamic and 24,632 Static
training windows.

\subsubsection{Scene Tokenizer and Decoder Training}

Multi-view pretraining uses 9,496 scenes, comprising 2,834 from DL3DV,
2,818 from TartanAir, 2,142 from RealEstate10K, 941 from CO3D, and 761
from WildRGB-D. To train the tokenizer with different visible-view
configurations, anchor-view and random-view dropout each use probability
0.4 at the memory level. For each training sample with at least two source views,
we apply one of three mutually exclusive memory-level operations: mask the
anchor-view patch block with probability 0.4, mask one uniformly sampled
non-anchor view block with probability 0.4, or retain all view blocks with
probability 0.2. Single-view samples retain their only view.

During Waymo adaptation, we use the Static subset described in
Section~\ref{app:Waymo Data Construction}. Since these windows contain no
visible moving tracks under our motion criterion, we allow temporal
source--query separations up to 2.0\,s while the observation and query
cameras move. This sampling covers both nearby and more distant query viewpoints, so that viewpoints far from the anchor remain within the training distribution.
The adaptation schedule uses a 2k-step warmup followed by cosine
learning-rate decay; training budgets, initial learning rates, and
effective batch sizes follow Section~\ref{sec:experimental_setup}.
For the geometry objective defined in Eq.~\ref{eq:representation_loss}, we set
the gradient-loss weight to $0.1$, yielding
\begin{equation}
\mathcal L_{\mathrm{rep}}
=
\mathcal L_{\log D}
+0.1\mathcal L_{\mathrm{grad}}.
\label{eq:app_geometry_loss}
\end{equation}

\subsubsection{World-Model Training}

World Model training uses 139,367 windows from 798 usable Waymo training
segments covering static and dynamic content. Each window contains four
context states and seven future states at 0.2\,s intervals, corresponding
to a forecast horizon of 1.4\,s from the last context state. 
For one-step initialization, each window supplies seven
$4\rightarrow1$ transitions with observation-encoded histories.

The concrete loss corresponding to Eq.~\ref{eq:init_loss} is
\begin{equation}
\mathcal L_{\mathrm{init}}
=
\operatorname{SmoothL1}_{0.1}(\widehat S,S^*)
+0.05\left[1-\cos\!\left(\operatorname{vec}\widehat S,
                              \operatorname{vec}S^*\right)\right]
+0.24\mathcal L_{\log D}.
\label{eq:app_one_step_loss}
\end{equation}
The SmoothL1 transition parameter is 0.1, and cosine similarity is
computed between flattened states. During autoregressive training,
Eq.~\ref{eq:rollout_loss} weights the FRONT log-depth losses equally
across horizons. Each window requires one GFM forward pass on its twelve
context images; future LiDAR depth supplies the reconstruction targets.

We use AdamW with $\beta=(0.9,0.95)$, weight decay $0.05$,
gradient clipping at $1.0$, and learning-rate warmup followed by
cosine decay. 

\subsection{Data and Evaluation Protocols}
\label{app:evaluation_protocol}
\subsubsection{Datasets and Temporal Sampling}
We construct one candidate window per sequence. Each window contains four
historical observations ending at anchor $t$ and ten future targets.
Eligibility is determined solely by the availability of the required RGB
images, camera poses, and LiDAR measurements, together with timestamp
matching. Using a fixed temporal position and at most
one window per sequence prevents longer sequences from receiving greater
weight.

For Waymo, the candidate pool contains one window from each of the 202
segments in the official validation split. We define logical states at
5\,Hz and place the anchor at the tenth logical state, 1.8\,s after the
first complete timestamp. A complete timestamp contains FRONT,
FRONT\_LEFT, and FRONT\_RIGHT images together with the vehicle pose. The
four observations occur at $t-0.6$, $t-0.4$, $t-0.2$, and $t$, and the ten
targets span $t+0.2$ through $t+2.0$\,s. For each nominal time, we select
the nearest complete timestamp within 20\,ms. We partition the 202 segments
into 102 segments used for checkpoint selection and 100 held-out segments
reserved for final evaluation. The evaluation subset is sampled uniformly
without replacement using seed 20260804, after which its segment identities
and anchors are fixed for all paired comparisons. No segment in the held-out
evaluation subset is used for checkpoint selection. This fixed sample
contains 79 Dynamic and 21 Static windows under the partition defined in
Appendix~\ref{app:Waymo Data Construction}.

For KITTI, we consider the 151 sequences covered by the KITTI Depth
Completion annotations and construct at most one fixed window per sequence. Logical states follow the VGGT-World
two-raw-frame stride, with the anchor placed at the tenth logical state,
corresponding to raw-frame offset 18 from the first image. We retain a drive
when all four historical and ten future RGB, pose, and LiDAR records are
available on this stride, yielding 113 eligible drives. Ineligible drives
are recorded as skipped and are not replaced. This common anchor placement
keeps the temporal position comparable to Waymo while respecting KITTI's
native sampling cadence.

For nuScenes, we construct one window for each of the 150 scenes in the
official validation split. Because CAM\_FRONT is sampled asynchronously,
the anchor is defined as the image nearest to 1.8\,s after the scene start,
subject to a 120\,ms tolerance, rather than as the tenth stored image.
Historical observations occur at $t-0.75,t-0.50,t-0.25,t$, and ten future
targets follow at 0.25\,s intervals. Each target is paired with the nearest
LIDAR\_TOP sweep within 50\,ms. This construction preserves a comparable
fixed temporal anchor across datasets while accommodating the native
nuScenes sensor timing. 

\subsubsection{Input and Query Construction}

Our method follows the balanced-resolution preprocessing rule of
VGGT-$\Omega$. It preserves the nominal patch-token budget while adapting the
height and width to the source aspect ratio and rounding both dimensions to
multiples of the 16-pixel patch size; a centered crop is applied only when
needed to limit extreme aspect ratios. The resulting evaluation rasters
(height$\times$width) are $416\times624$ for Waymo, $368\times720$ for KITTI,
and $384\times688$ for nuScenes. Camera intrinsics are updated using the same
resize factors and crop offsets. Since VGGT-$\Omega$ assumes a centered principal point,
its predicted intrinsics can introduce systematic errors
in viewing-ray directions when the actual principal point
is off-center. We therefore use dataset-calibrated
intrinsics together with VGGT-$\Omega$-estimated camera
poses for source-view geometric conditioning. External baselines retain their respective
preprocessing conventions.

At each Waymo observation time, we use synchronized FRONT, FRONT\_LEFT, and
FRONT\_RIGHT images. VGGT-$\Omega$ jointly processes all twelve historical
images, placing the latest FRONT image first to define a shared reference
frame. A timestamp-specific memory mask exposes only the three view blocks
belonging to each state to the Scene Tokenizer. The four states are then supplied to
the World Model in chronological order. KITTI and nuScenes use one camera
view per timestamp.

The World Model predicts ten successive states with a four-state sliding
window. GT-camera and predicted-camera evaluations share this state
trajectory and differ in the decoder query poses. The camera predictor
receives four observed metric poses and the current forward-facing image,
as detailed in Appendix~\ref{app:camera_predictor}. Within each camera setting,
Copy-Last and forecasting share query cameras and LiDAR evaluation masks.

Camera-matched Copy-Last holds $S_t$ fixed and evaluates
$G_\psi(S_t,Q_{t+h})$, retaining the representation, decoder, and query
trajectory used by forecasting. Fixed-View Copy-Last repeats the raw
VGGT-$\Omega$ FRONT depth map from the observed anchor at every future
timestamp. The map remains fixed while the future LiDAR reference changes.

\subsubsection{Depth, Scale, and Metric Computation}

Depth is evaluated against projected LiDAR measurements in the query-camera
frame. For a transformed point $(x_c,y_c,z_c)$, the reference depth is $z_c$.
The valid set $\Omega$ contains projected measurements with finite depth
satisfying $0<D\leq80$\,m. Predictions and references are compared at corresponding
pixels of the evaluation raster:
\begin{align}
\operatorname{AbsRel}(\widehat D,D)
&=\frac{1}{|\Omega|}\sum_{p\in\Omega}
  \frac{|\widehat D(p)-D(p)|}{D(p)},\\
\delta_1(\widehat D,D)
&=\frac{1}{|\Omega|}\sum_{p\in\Omega}
  \mathbf{1}\!\left[
  \max\!\left(\frac{\widehat D(p)}{D(p)},
              \frac{D(p)}{\widehat D(p)}\right)<1.25
  \right].
\end{align}
For all depth-forecasting results, we compute each metric independently on
every target frame and then average over frames, giving every frame equal
weight regardless of its number of valid LiDAR pixels.

Our forecasts calibrate metric scale once from same-pixel
VGGT-$\Omega$/LiDAR depth pairs at the observed anchor. The conversion and
scene-normalization unit defined in Appendix~\ref{app:scale_protocol} are
shared by the input states, query-camera translations, and decoded depths
throughout the rollout. Future LiDAR provides the scoring reference, while
decoded predictions retain the context-derived scale. Fixed-View Copy-Last
uses the same-pixel anchor calibration for its raw GFM depth raster.

For VGGT-World, we follow its original per-target-frame scale-alignment
protocol. FR3D numbers are quoted from its original evaluation, which uses
per-target-frame scale-and-shift alignment. DINO-Foresight relative
inverse-depth predictions are fitted to target-frame LiDAR for depth scoring.

\subsubsection{Compressed VGGT State Baseline}
\label{app:compressor}

\paragraph{Architecture.}
The representation ablation uses an autoencoder of
VGGT-$\Omega$ patch features. For each training example, the frozen GFM
jointly processes four historical timestamps and one target timestamp, with
FRONT, FRONT\_LEFT, and FRONT\_RIGHT at each timestamp. We retain the target
FRONT features from GFM layers $\{4,11,17,23\}$; each layer provides a
$26\times39$ grid of 2,048-dimensional patch features. Four independent
linear projections reduce each feature vector from 2,048 to 128 dimensions.
The projected features are concatenated channel-wise and spatially
average-pooled from $26\times39$ to $16\times32$, producing a compressed state
$Z\in\mathbb{R}^{512\times512}$. To reconstruct the GFM features, the state is
bilinearly upsampled to $26\times39$, split into four 128-dimensional groups,
and mapped back to four 2,048-dimensional feature grids by independent linear
layers. A frozen VGGT-$\Omega$ DPT head decodes the reconstructed features into
FRONT depth. The compressor has 2.11M trainable parameters.

\paragraph{Training.}
We follow the same two-stage training paradigm as the Scene Tokenizer. The
compressor is first pretrained for 200k steps on the same 9,496-scene
multi-view dataset, after which the best checkpoint is selected and further
fine-tuned on Waymo for 50k steps.

The training procedure is identical in both stages. For each training sample,
we extract the VGGT-$\Omega$ features of the target image and compress them
into the latent state $Z$. The compressor then reconstructs the original
multi-layer VGGT-$\Omega$ features, which are passed through the frozen
VGGT-$\Omega$ DPT head to obtain the predicted depth. The prediction is
supervised directly against the corresponding ground-truth depth using
\begin{equation}
\mathcal L_{\mathrm{rep}}
=
\mathcal L_{\log D}
+0.1\mathcal L_{\mathrm{grad}}.
\end{equation}
The VGGT-$\Omega$ backbone and DPT head remain frozen throughout both stages,
and gradients are propagated through the DPT head to optimize only the
compressor.

\subsection{Cross-Query Geometric Consistency}
\label{app:query_consistency}

\subsubsection{Forecast and Observation-Encoded States}

We use the same 100 Waymo validation windows and all ten future
timestamps. World denotes the world model autoregressive prediction.
Enc.\ denotes an independent target-time reconstruction from exactly
three images. For Enc., a separate historical pass provides the normalization length
$L_0$ defined in Eq.~\ref{eq:app_physical_unit}.

For Enc., raw target FRONT GFM depth is calibrated using target-time
LiDAR, and the normalization scale is set to its paired scale times
$L_0$. World retains its historical-anchor calibration.
Both decode under the same GT FRONT reference query and the same
camera perturbations, without independently aligning decoded depths.
Enc.\ is therefore an observation-based reference with privileged
target-time scale information.

\subsubsection{Camera Perturbations and Coverage}

Each perturbation combines a 3D translation with an axis--angle
rotation. Translation directions and rotation axes are independently
sampled uniformly on the unit sphere, while their magnitudes are sampled
uniformly within one of three perturbation levels. Small perturbations
use translation magnitudes in $[0,0.5]$\,m and rotation magnitudes in
$[0,2]^\circ$; Medium uses $[0.5,1.0]$\,m and $[2,5]^\circ$; and Large
uses $[1.0,1.5]$\,m and $[5,10]^\circ$. Rotation magnitudes refer to the
total axis--angle rotation, and translation magnitudes to Euclidean
displacement.
For a reference camera-to-world pose $P_0$, the perturbed pose is
$P_i=P_0\Delta_i$. Perturbations are expressed in the reference
camera's local coordinates, with $+x$ right, $+y$ down, and $+z$
forward.

Using seed 20260804, we initially sample 250 perturbations at each level.
For each perturbation, we compute its mean bidirectional coverage across
all evaluation windows, timestamps, and both state variants, and retain
the 200 perturbations with the highest coverage within each level. This
produces a fixed set of 600 perturbations from 750 initial samples.
The target-three-view evaluation reuses this fixed subset.

\subsubsection{Reprojection, Masks, and Aggregation}

We back-project $D_a$ using $K_a^{-1}$, transform points into camera
$b$, and project them using $K_b$. Destination pixels are rounded
to their nearest integer location. A z-buffer retains the smallest
positive destination depth when points collide. Out-of-frame
projections and points with destination $z\leq10^{-4}$\,m are
excluded. Let $\Omega_{a\rightarrow b}$ contain pixels with finite,
positive reprojected and decoded destination depths. The directional
error is
\begin{equation}
e_{a\rightarrow b}
=
\frac{1}{|\Omega_{a\rightarrow b}|}
\sum_{p\in\Omega_{a\rightarrow b}}
\frac{|\mathcal R_{a\rightarrow b}(D_a)(p)-D_b(p)|}
     {\max(D_b(p),10^{-8})}.
\end{equation}
We report $(e_{0\rightarrow i}+e_{i\rightarrow0})/2$.
This dense consistency comparison uses neither the sparse LiDAR
evaluation mask nor the 80\,m depth cutoff. No additional
depth-agreement threshold filters the overlapping pixels.

Coverage is
$(|\Omega_{0\rightarrow i}|+|\Omega_{i\rightarrow0}|)/(2HW)$.
Each horizon equally averages 100 windows and 200 perturbations
per level; the overall mean averages all ten horizons.
There are $100\times10\times600=600{,}000$ query pairs per method.
This metric measures geometric agreement across queries and complements
the evaluation of accuracy against physical ground truth.

\subsection{Feature Analysis of VGGT-\texorpdfstring{$\Omega$}{Omega}}
\label{app:gfm_variation}
VGGT-$\Omega$ jointly processes multiple images, so the representation of an
image may depend on the other images included in the same forward pass. This
property is relevant to state supervision: an observation-derived target can
encode both scene geometry and context-specific variation introduced by the
GFM. We therefore examine whether the features of an unchanged image remain
stable when only its surrounding context changes.

We conduct a controlled diagnostic on all 100 Waymo validation windows. For
each window, the same FRONT image at the first future timestamp
($t{+}0.2$\,s) is retained at the same input index, with its image content,
camera, and reference frame fixed. The reference frame is defined by the
FRONT image at the observed anchor $t$. Each timestamp contributes three views. The shortest prefix containing the probed image comprises
four observed timestamps and the first future timestamp, giving
$5\times3=15$ images. The complete diagnostic window contains the same four
observed timestamps and seven future timestamps, giving
$11\times3=33$ images. We compare the probed-image features produced under
these two contexts. For feature layer $\ell$, we measure Relative L2
\begin{equation}
E_F^{(\ell)}
=
\frac{\lVert F_{33}^{(\ell)}-F_{15}^{(\ell)}\rVert_2}
     {\lVert F_{15}^{(\ell)}\rVert_2}.
\label{eq:app_gfm_context_feature}
\end{equation}
We additionally compare the dense DPT depth maps predicted for the same
$t{+}0.2$\,s FRONT image under both unaligned and calibrated settings.
The unaligned comparison is performed directly on the raw DPT outputs.
For the calibrated comparison, both forwards estimate their
VGGT-$\Omega$-to-metric factors from the same matched
VGGT-$\Omega$/LiDAR depths at the observed anchor $t$,
following the physical-unit convention in
Eq.~\ref{eq:app_physical_unit}.

\begin{table}[t]
\centering
\caption{\textbf{Context dependence of frozen VGGT-$\Omega$.} All metrics come from the same image while varying only its jointly
processed context. Values are sample mean $\pm$ standard deviation.}
\label{tab:app_context_diagnostic}
\small
\setlength{\tabcolsep}{6pt}
\renewcommand{\arraystretch}{1.05}
\begin{tabular}{@{}llc@{}}
\toprule
\textbf{Quantity} & \textbf{Measure} & \textbf{15 vs. 33 images}\\
\midrule
\multicolumn{3}{@{}l}{\textit{GFM patch features}}\\
Layer-4 patch features & Relative L2 & $1.38\pm0.59\%$\\
Layer-11 patch features & Relative L2 & $11.21\pm1.96\%$\\
Layer-17 patch features & Relative L2 & $33.55\pm8.71\%$\\
Layer-23 patch features & Relative L2 & $31.15\pm5.97\%$\\
\textbf{Four-layer mean} & Relative L2 & $\mathbf{19.32\pm4.16\%}$\\
\midrule
\multicolumn{3}{@{}l}{\textit{DPT depth of the same image}}\\
Raw DPT depth & AbsRel & $12.08\pm7.59\%$\\
Anchor-calibrated DPT depth & AbsRel & $\mathbf{3.82\pm1.58\%}$\\
\bottomrule
\end{tabular}
\end{table}

As shown in Table~\ref{tab:app_context_diagnostic}, context dependence is
small in the early representation but becomes pronounced in deeper layers.
Relative feature differences increase from $1.38\%$ at layer 4 to
$33.55\%$ and $31.15\%$ at layers 17 and 23, respectively, with a
four-layer mean of $19.32\%$. The raw DPT outputs differ by $12.08\%$
AbsRel. After expressing both outputs in the anchor-calibrated metric unit,
the difference decreases to $3.82\%$. Although feature relative L2 and depth
AbsRel are not directly commensurate, this controlled contrast shows that
substantial variation in the VGGT-$\Omega$ feature coordinates can coexist
with considerably more stable metric geometry. The features therefore carry
context-conditioned information that is not uniquely determined by scene
geometry, in addition to the geometric information used by the DPT head.

This property inevitably affects a state representation built on
VGGT-$\Omega$ features. The Scene Tokenizer can suppress part of the
context-specific variation, but its observation-derived targets are still
constructed from GFM features that are not purely geometric. This observation
helps explain the state-supervision ablation in
Table~\ref{tab:ablation_supervision}. During one-step initialization, latent
supervision provides a useful local target for learning the state transition.
During autoregressive rollout, however, exact latent matching then asks the World Model
to reproduce both scene geometry and context-specific feature variation,
including directions with limited influence on scene geometry.
Supervising rollout states through decoded depth instead constrains the
scene geometry required at inference while allowing flexibility in these
context-sensitive latent directions. This analysis supports our use of
latent supervision for one-step initialization and decoded-geometry
supervision for closed-loop rollout.

\subsection{Additional Quantitative Results}
\label{app:additional_results}

\subsubsection{Single-View-Controlled Baseline Comparison}
\label{app:Single-View-Controlled Baseline Comparison}

The main GeoState evaluation uses the synchronized FRONT, FRONT\_LEFT, and
FRONT\_RIGHT images at each observed timestamp, whereas our compared methods use a single
image. To control for this difference in input-view count, we re-evaluate
GeoState on the same 100 Waymo evaluation windows using only FRONT. Both the frozen GFM
and the Scene Tokenizer therefore receive four historical images in total,
one from each observed timestamp. The model checkpoints, timestamps, query
cameras, metric-scale calibration, and LiDAR masks remain unchanged.
Table~\ref{tab:app_front_only_forecasting} compares this FRONT-only variant
with the external forecasting baselines from Table~\ref{tab:main_waymo}.


\begin{table}[t]
\centering
\caption{\textbf{Long-horizon depth forecasting on Waymo with FRONT-only
history.} GeoState receives only the FRONT image at each observed timestamp;
Bold and underlined values indicate the best and second-best results among
methods using predicted future cameras.}
\label{tab:app_front_only_forecasting}

\begingroup
\footnotesize
\setlength{\tabcolsep}{2.3pt}
\renewcommand{\arraystretch}{1.02}
\setlength{\aboverulesep}{1.5pt}
\setlength{\belowrulesep}{1.5pt}
\setlength{\cmidrulesep}{1pt}

\begin{adjustbox}{max width=\linewidth}
\begin{tabular}{@{}l c c *{4}{cc}@{}}
\toprule
\textbf{Method}
& \textbf{Camera}
& \textbf{Per-frame}
& \multicolumn{2}{c}{$t{+}0.2$ s}
& \multicolumn{2}{c}{$t{+}0.6$ s}
& \multicolumn{2}{c}{$t{+}1.0$ s}
& \multicolumn{2}{c}{$t{+}2.0$ s} \\
\cmidrule(lr){4-5}
\cmidrule(lr){6-7}
\cmidrule(lr){8-9}
\cmidrule(l){10-11}
& & \textbf{alignment}
& AbsRel $\downarrow$ & $\delta_1 \uparrow$
& AbsRel $\downarrow$ & $\delta_1 \uparrow$
& AbsRel $\downarrow$ & $\delta_1 \uparrow$
& AbsRel $\downarrow$ & $\delta_1 \uparrow$ \\
\midrule

DINO-Foresight & Pred. & Yes
& 0.119 & 0.841
& 0.138 & 0.825
& 0.151 & 0.798
& 0.243 & 0.663 \\

VGGT-World & Pred. & Yes
& \underline{0.073} & \underline{0.931}
& \underline{0.112} & \underline{0.892}
& 0.143 & 0.826
& 0.213 & 0.731 \\

FR3D & Pred. & Yes
& 0.110 & 0.879
& 0.121 & 0.858
& \underline{0.131} & \underline{0.840}
& \underline{0.152} & \underline{0.800} \\

\rowcolor{oursblue}
GeoState & Pred. & No
& \textbf{0.070} & \textbf{0.936}
& \textbf{0.089} & \textbf{0.914}
& \textbf{0.105} & \textbf{0.881}
& \textbf{0.143} & \textbf{0.813} \\

\bottomrule
\end{tabular}
\end{adjustbox}
\endgroup
\end{table}

Under predicted future cameras, FRONT-only GeoState outperforms FR3D,
VGGT-World, and DINO-Foresight in both AbsRel and $\delta_1$ at every reported
horizon. At $t{+}2.0$\,s, it obtains an AbsRel of 0.143 and a $\delta_1$ of
0.813, compared with 0.152/0.800 for FR3D, 0.213/0.731 for VGGT-World, and
0.243/0.663 for DINO-Foresight. This comparison removes the additional
side-view observations available to the main GeoState configuration, while
GeoState continues to use a single context-derived scale rather than
per-target-frame alignment. The performance advantage over the external
forecasting baselines therefore cannot be attributed solely to using three
camera views per observed timestamp.

\subsubsection{Motion-Stratified Forecasting Results}
\label{app:Motion-Stratified Forecasting Results}
We further report results under the Dynamic/Static partition defined in
Appendix~\ref{app:Waymo Data Construction}. Dynamic combines all windows containing
at least one qualifying visible moving track, while Static contains the
remaining windows. Table~\ref{tab:app_motion_stratified_forecasting} averages
each subset over all ten forecasting horizons and compares GeoState with the
camera-matched Copy-Last reference.

\begin{table}[t]
\centering
\caption{\textbf{Motion-stratified forecasting on Waymo.}
Results are frame-wise means over all ten horizons on the Waymo validation
set. ``Improve'' is the relative AbsRel reduction of GeoState over the
camera-matched Copy-Last, and $\Delta\delta_1$ is the absolute accuracy gain.}
\label{tab:app_motion_stratified_forecasting}

\begingroup
\footnotesize
\setlength{\tabcolsep}{4.0pt}
\renewcommand{\arraystretch}{1.05}

\begin{adjustbox}{max width=\linewidth}
\begin{tabular}{@{}llcccccc@{}}
\toprule
& & \multicolumn{3}{c}{\textbf{AbsRel}$\downarrow$}
& \multicolumn{3}{c}{\boldmath$\boldsymbol{\delta_1}\uparrow$\unboldmath} \\
\cmidrule(lr){3-5}\cmidrule(l){6-8}
\textbf{Subset} & \textbf{Camera}
& Copy-Last & GeoState & Improve
& Copy-Last & GeoState & $\Delta\delta_1$ \\
\midrule
Dynamic & GT
& 0.0980 & 0.0852 & 13.10\%
& 0.8842 & 0.9077 & +0.0236 \\
Dynamic & Pred.
& 0.1070 & 0.0967 & 9.63\%
& 0.8717 & 0.8946 & +0.0229 \\
\addlinespace[2pt]
Static & GT
& 0.0710 & 0.0675 & 4.92\%
& 0.9245 & 0.9337 & +0.0092 \\
Static & Pred.
& 0.0915 & 0.0895 & 2.17\%
& 0.8917 & 0.8995 & +0.0078 \\
\bottomrule
\end{tabular}
\end{adjustbox}
\endgroup
\end{table}

GeoState improves both subsets under GT and predicted query cameras. On the
Dynamic subset, it reduces AbsRel by 13.10\% with GT cameras and 9.63\% with
predicted cameras, with corresponding $\delta_1$ gains of 0.0236 and 0.0229.
The gains on Static windows are smaller but remain positive: 4.92\% and
2.17\% relative AbsRel reductions, respectively. The substantially larger
gains on Dynamic windows indicate that the World Model effectively captures
temporal changes in scene geometry. The same pattern under GT query
cameras further shows that this dynamic-scene advantage is present
independently of future-camera prediction error.
\end{document}